\documentclass[10pt,twocolumn,letterpaper]{article}
\usepackage[T1]{fontenc}
\usepackage{amsmath}
\usepackage{amsthm}
\usepackage{amssymb}
\usepackage{newtxtext}
\usepackage{helvet}
\usepackage{courier}
\usepackage[hyphens]{url}
\usepackage{graphicx}
\usepackage[round,authoryear]{natbib}
\usepackage{caption}
\usepackage{placeins}
\usepackage{xcolor}
\usepackage{booktabs}
\usepackage{colortbl}
\usepackage{tikz}
\usetikzlibrary{arrows.meta,positioning,calc,backgrounds,fit,shadows}
\usepackage{xspace}
\usepackage{cleveref}
\usepackage{fontawesome5}
\makeatletter
\renewenvironment{abstract}{%
  \centerline{\bfseries Abstract}%
  \vspace{0.5ex}%
  \setlength{\leftmargini}{10pt}%
  \begin{quote}\small
}{%
  \par\end{quote}\vskip 1ex
}
\renewcommand\section{\@startsection{section}{1}{\z@}%
  {-2.0ex plus -0.5ex minus -.2ex}{3pt plus 2pt minus 1pt}%
  {\Large\bfseries\centering}}
\renewcommand\subsection{\@startsection{subsection}{2}{\z@}%
  {-2.0ex plus -0.5ex minus -.2ex}{3pt plus 2pt minus 1pt}%
  {\large\bfseries\raggedright}}
\renewcommand\subsubsection{\@startsection{subparagraph}{3}{\z@}%
  {-6pt plus -2pt minus -1pt}{-1em}{\normalsize\bfseries}}
\renewcommand\paragraph{\@startsection{paragraph}{4}{\z@}%
  {-6pt plus -2pt minus -1pt}{-1em}{\normalsize\bfseries}}
\renewcommand\normalsize{\@setfontsize\normalsize\@xpt{11}}
\renewcommand\small{\@setfontsize\small\@ixpt{10}}
\renewcommand\footnotesize{\@setfontsize\footnotesize\@ixpt{10}}
\renewcommand\scriptsize{\@setfontsize\scriptsize\@viipt{10}}
\renewcommand\large{\@setfontsize\large\@xipt{12}}
\renewcommand\Large{\@setfontsize\Large\@xiipt{14}}
\renewcommand\LARGE{\@setfontsize\LARGE\@xivpt{16}}
\makeatother

\setcitestyle{aysep={}}
\crefname{equation}{Eq.}{Eqs.}
\Crefname{equation}{Eq.}{Eqs.}
\crefname{figure}{Fig.}{Figs.}
\Crefname{figure}{Fig.}{Figs.}

\newcounter{gap}
\newcommand{\gapheading}[1]{\refstepcounter{gap}\paragraph{Gap~\thegap: #1}}
\crefname{gap}{Gap}{Gaps}
\Crefname{gap}{Gap}{Gaps}

\newcommand{\R}{\mathbb{R}}
\newcommand{\E}{\mathbb{E}}

\newcommand{\latent}{z}                      \newcommand{\goal}{z_g}                       \newcommand{\traj}{\tau}                       \newcommand{\trajp}{\tau^{+}}                  \newcommand{\trajn}{\tau^{-}}                  \newcommand{\zhat}{\hat{z}}                                                  \newcommand{\trajhat}{\hat{\tau}}              \newcommand{\Spsi}{S_{\psi}}                   \newcommand{\enc}{\mathcal{E}_{\theta}}         \newcommand{\pred}{\mathcal{F}_{\phi}}

\newcommand{\Lpred}{\mathcal{L}_{\mathrm{pred}}}
\newcommand{\Lsig}{\mathcal{R}_{\mathrm{sigreg}}}
\newcommand{\Lpath}{\mathcal{L}_{\mathrm{path}}}
\newcommand{\Lmined}{\mathcal{L}_{\mathrm{mined}}}
\newcommand{\Lvalewm}{\mathcal{L}_{\mathrm{Traj\text{-}LeWM}}}

\newcommand{\Czero}{C_{0}}

\newcommand{\Clambda}{C_{\lambda}}
\newcommand{\Stilde}{\widetilde{S}_{\psi}}

\definecolor{PaperCream}{HTML}{F1E6D8}
\definecolor{PaperRed}{HTML}{AA2B3A}
\definecolor{PaperNavy}{HTML}{28314E}

\newcommand{\method}{Traj-LeWM\xspace}
\newcommand{\methodsyn}{Traj-LeWM-syn\xspace}

\newcommand{\overviewfigure}{\begin{figure*}[!t]
\centering
\includegraphics[width=\textwidth]{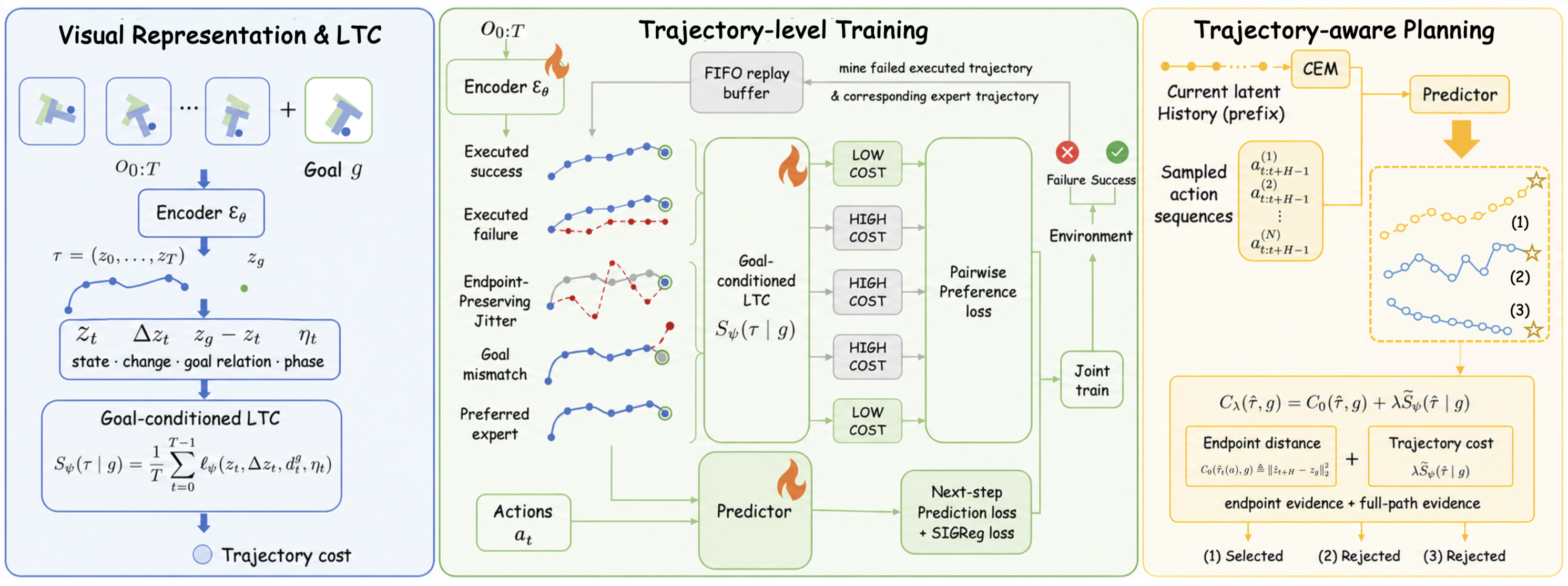}
\caption{Overview of \method. Stage 1 encodes the observation trajectory
and goal into latent representations used by LTC. Stage 2 trains LTC
with synthetic and mined closed-loop trajectory preferences while
retaining LeWM's next-step prediction objective. Stage 3 ranks CEM
candidates using the joint endpoint-plus-LTC score.}
\label{fig:overview}
\end{figure*}}

\newtheorem{definition}{Definition}
\theoremstyle{remark}

\title{Traj-LeWM: Path-Aware World-Model Planning via Latent Trajectory Cost}
\date{}
\author{
Xiaodi Huang\textsuperscript{\rm 1,2\,*}\quad
Ziyi Ding\textsuperscript{\rm 3,5\,*}\quad
Jingtian Wan\textsuperscript{\rm 6,5}\quad
Yuchen Liu\textsuperscript{\rm 4}\quad
Yuan Zhang\textsuperscript{\rm 7}\\
Xiao-Ping Zhang\textsuperscript{\rm 3}\quad
Jiayu Chen\textsuperscript{\rm 4,5\,\textdagger}\quad
Zhang Zhang\textsuperscript{\rm 1\,\textdagger}\quad
Tao Huang\textsuperscript{\rm 2\,\textdagger}\\[0.35em]
{\normalfont\normalsize\faGithub\ \url{https://github.com/XiaodiHuang-code/Traj_LeWM}}
}

\makeatletter
\renewcommand{\maketitle}{%
  \twocolumn[\vbox to 2.25in{%
    \hsize\textwidth
    \linewidth\hsize
    \vskip 0.625in minus 0.125in
    \centering
    {\LARGE\bfseries\@title\par}
    \vskip 0.1in plus 0.5fil minus 0.05in
    {\Large\bfseries\@author\par}
    \vskip 1em plus 2fil
  }]%
  \thispagestyle{empty}%
}
\makeatother

\newcommand{\TrajLeWMArxivFootnote}{\begingroup
  \renewcommand{\thefootnote}{}\footnotetext{\raggedright
    \textsuperscript{\rm *}Equal contribution.
    \textsuperscript{\textdagger}Corresponding authors.\newline
    \textsuperscript{\rm 1}Institute of Automation, Chinese Academy of Sciences;
    \textsuperscript{\rm 2}Shanghai Jiao Tong University;
    \textsuperscript{\rm 3}Tsinghua Shenzhen International Graduate School;
    \textsuperscript{\rm 4}The University of Hong Kong;
    \textsuperscript{\rm 5}INFIFORCE;
    \textsuperscript{\rm 6}University of Science and Technology of China;
    \textsuperscript{\rm 7}Peking University.}\endgroup
}

\begin{document}
\maketitle
\TrajLeWMArxivFootnote

\begin{abstract}
LeWM is a lightweight visual world model that learns latent dynamics
end-to-end from pixels and ranks candidate action sequences by the
distance between their predicted endpoints and the goal. However, LeWM
has two limitations. First, during training, it learns local next-step
transitions without evaluating complete trajectories relative to the
task goal. Second, during planning, it ranks candidates solely by
predicted endpoint distance. Because model predictions may differ from
actual execution outcomes, the candidate whose predicted endpoint is
closest to the goal may not perform best when executed in the
environment. The evolution of the complete predicted trajectory can
therefore provide complementary information beyond endpoint distance.
To address these limitations, we propose \method, which retains LeWM's
local-dynamics objective and endpoint score while introducing a
goal-conditioned latent trajectory cost (LTC) that aggregates
trajectory-level information as a complementary signal. During training,
LTC-based trajectory-preference supervision complements next-step
prediction in shaping the shared representation. During planning, LTC is
combined with endpoint distance to incorporate intermediate-path
information into candidate ranking.
With joint endpoint-plus-LTC scoring, \method outperforms LeWM on Push-T,
OGBench-Cube, Reacher, and Two-Room by $3$, $14$, $7$, and $7$ percentage
points, respectively. Controlled experiments and ablations further
verify the complementary roles of trajectory-level representation
shaping and path-aware candidate ranking.
\end{abstract}

\section{Introduction}
\label{sec:intro}

\overviewfigure

Joint-embedding predictive architectures (JEPAs) support model-based
control by predicting future observations in compact latent spaces rather
than reconstructing decision-irrelevant pixel details
~\citep{lecun2022path,assran2023ijepa}. LeWM
~\citep{maes_lelidec2026lewm} extends this paradigm into a lightweight
visual world model trained end-to-end from pixels and achieves strong
performance in goal-conditioned planning. However, this performance does
not remain consistent on more challenging tasks: LeWM's success rate
drops from $96\%$ on Push-T to $74\%$ on OGBench-Cube, whereas DINO-WM
reports $86\%$ success on Cube~\citep{zhou2024dinowm}. Across tasks, we empirically find that candidates with the same start and
goal and similar endpoint costs can produce different execution outcomes,
suggesting that endpoint information alone cannot reliably capture
candidate quality. This motivates using complete-trajectory information
in both representation learning and candidate ranking.

We attribute this behavior to two missing uses of trajectory information.
First, LeWM's next-step prediction loss supervises local transitions
between adjacent states, while SIGReg only regularizes the distribution
of latent representations; neither requires the shared encoder to
preserve information about a complete trajectory relative to its task
goal. Consequently, although autoregressive rollout produces a sequence
of predicted states, the representation space has not been directly
trained to assess the goal-conditioned quality of a complete candidate
trajectory. Second, LeWM ranks candidates only by the distance between
their predicted latent endpoints and the goal representation. A predicted
endpoint close to the goal does not guarantee that the same action
sequence will perform well when executed in the environment, and
candidates with similar predicted endpoints cannot be distinguished
using their different intermediate evolutions.

To address these limitations, we propose \method. It retains LeWM's
original prediction objective and endpoint score while introducing a
goal-conditioned latent trajectory cost (LTC) over complete latent
trajectories. The overall framework is illustrated in
\Cref{fig:overview}. LTC maps the goal-relative evolution of a complete
latent trajectory to a learned scalar cost. During training, LTC is
learned from pairwise preferences that favor goal-matched expert
trajectories over synthetic negative trajectories and closed-loop
execution failures; synthetic preferences also shape the shared encoder.
During planning, LTC is combined with endpoint distance so that candidate
ranking uses both endpoint and intermediate-path information. Across four
simulated tasks, \method consistently outperforms LeWM, while evaluation
on a physical Franka FR3 further demonstrates its real-robot feasibility.
Controlled experiments separately show that trajectory supervision
improves the shared representation and that LTC uses intermediate-path
information to improve candidate ranking.

In summary, our contributions are threefold:
\begin{itemize}
    \item We introduce a learned goal-conditioned latent trajectory cost
    that evaluates complete predicted trajectories. By aggregating
    goal-relative latent evolution over time, LTC complements endpoint
    distance with a path-sensitive planning signal.

    \item We develop a trajectory-preference learning mechanism based on
    synthetic negative trajectories and closed-loop execution failures.
    Synthetic preferences introduce trajectory-level, goal-conditioned
    supervision into the shared representation, while mined failures
    further train LTC without changing LeWM's original predictor
    objective.

    \item We conduct a comprehensive evaluation on four simulated tasks
    and a physical Franka FR3. Compared with LeWM, \method improves
    success rates by $3$, $14$, $7$, and $7$ percentage points on Push-T,
    OGBench-Cube, Reacher, and Two-Room, respectively, and increases
    real-robot success from $50\%$ to $70\%$ over the same $20$ tasks.
    Controlled endpoint-only, endpoint-matched, and fixed-endpoint
    analyses, together with ablations, further verify the effects of
    trajectory-level representation shaping and intermediate-path-aware
    candidate ranking.
\end{itemize}

\section{Related Work}
\label{sec:related}

\paragraph{World models and JEPA-based planning.}
World models support decision-making by learning predictive dynamics and planning through imagined rollouts. Representative approaches include generative latent agents such as Dreamer~\citep{hafner2020dreamer,hafner2023dreamerv3}, value-guided planners such as TD-MPC~\citep{hansen2022tdmpc,hansen2024tdmpc2}, and reward-free latent planners such as PLDM and DINO-WM~\citep{sobal2025learning,zhou2024dinowm}. Universal Planning Networks jointly learn representations and differentiable latent-space planning for image-conditioned control~\citep{srinivas2018upn}. Joint-embedding predictive architectures (JEPAs) instead predict masked or future content directly in representation space~\citep{assran2023ijepa,lecun2022path}. Recent systems extend this principle to action-conditioned physical planning: V-JEPA~2-AC plans from image goals on robots~\citep{assran2025vjepa2}, while a systematic study of JEPA world models analyzes how representation, prediction, and planner design affect physical planning~\citep{terver2026jepawm}. LeWM~\citep{maes_lelidec2026lewm} learns its encoder and predictor end-to-end from pixels, then ranks action sequences by predicted endpoint distance. Value-guided JEPA planning further shapes embedding distances to approximate a goal-conditioned value function~\citep{destrade2026valueguided}. Our method instead learns a cost over complete goal-conditioned latent rollouts and uses it alongside, rather than as, the endpoint distance.

\paragraph{Trajectory-level planning signals.}
Planning signals beyond endpoint distance include goal-conditioned value functions and reachability estimates~\citep{eysenbach2019sorb,kaelbling1993goals}, as well as visual representations trained to provide dense goal-conditioned rewards~\citep{ma2023vip}. Learned reward predictions support latent imagination in agents such as DreamerV3~\citep{hafner2023dreamerv3}, whereas optimal cost design learns surrogate objectives tailored to the finite horizon and replanning behavior of model predictive control~\citep{jain2021optimalcost}. A separate line treats trajectories as generated objects: denoising diffusion models provide the underlying generative machinery~\citep{ho2020denoising}, Diffuser adapts it to classifier-guided trajectory planning~\citep{janner2022diffuser}, and energy-based models define scalar landscapes over structured inputs~\citep{du2019ibm}. Goal-conditioned behavior cloning and offline value-learning methods, including GCBC, GCIVL, and GCIQL, instead learn policies or values directly from offline data~\citep{park2024ogbench,kostrikov2022iql}. LTC does not replace LeWM's dynamics model or candidate generator. It scores the same CEM candidates from their complete predicted latent paths and supplements the endpoint score during ranking.

\paragraph{Preference-based cost learning.}
Pairwise preference models learn scalar rankings from comparisons, with the Bradley--Terry model providing the standard logistic form~\citep{bradley1952rank}; the same principle underlies reward learning from human comparisons~\citep{christiano2017preferences}. At the trajectory level, T-REX learns a reward from ranked demonstrations~\citep{brown2019trex}, and D-REX automatically constructs rankings by perturbing a behavior-cloned policy to generate demonstrations of different quality~\citep{brown2020drex}. Energy-based models likewise represent relative plausibility through a learned scalar landscape~\citep{lecun2006tutorial,du2019ibm}. We use the standard pairwise objective rather than proposing a new preference loss. Our preferences act on complete goal-conditioned latent trajectories: negatives combine goal mismatches and endpoint-preserving perturbations with failures mined from closed-loop execution, and the learned cost both shapes the shared encoder during training and ranks predicted trajectories during planning.

 \section{Method}
\label{sec:method}

\paragraph{Overview.}
\method extends LeWM with a goal-conditioned latent trajectory cost (LTC), $\Spsi(\traj\mid g)$, that evaluates complete latent trajectories. As illustrated in \Cref{fig:overview}, trajectory-level preferences train LTC and shape the shared representation, while during planning LTC supplements endpoint distance with a path-sensitive signal for ranking candidate action sequences.

\subsection{Motivation: Two Gaps in LeWM}
\label{sec:motivation}

\paragraph{LeWM training and planning.}
The agent receives pixel observations $o_t\in\mathcal O$ and executes actions $a_t\in\mathcal A$. Following LeWM~\citep{maes_lelidec2026lewm}, an encoder $\enc:\mathcal O\rightarrow\R^d$ maps each observation to a latent state $\latent_t=\enc(o_t)$, and an action-conditioned predictor $\pred$ predicts the next latent state. We write the goal observation as $g$ and its representation as $\goal=\enc(g)$. LeWM trains the encoder and predictor with a next-step latent prediction loss $\Lpred$ and uses SIGReg, denoted by $\Lsig$, to prevent representation collapse:
\begin{equation}
\label{eq:ldyn}
\mathcal L_{\mathrm{LeWM}}
\triangleq \Lpred+\lambda_{\mathrm{sig}}\Lsig.
\end{equation}
At inference time, LeWM freezes the world model and autoregressively rolls out each candidate action sequence $a_{t:t+H-1}$. With a real-prefix length $L$, the context-augmented predicted latent trajectory is
\begin{equation}
\label{eq:context-rollout}
\trajhat_t(a)
=\big(\latent_{t-L+1:t},\zhat_{t+1:t+H}\big).
\end{equation}
LeWM scores a candidate solely by the distance between its predicted endpoint and the goal representation:
\begin{equation}
\label{eq:c0}
\Czero(\trajhat_t(a),g)
\triangleq\big\|\zhat_{t+H}-\goal\big\|_2^2.
\end{equation}
It then uses CEM to minimize \Cref{eq:c0} and executes the selected action sequence in a receding-horizon manner. We omit the current-time subscript below and write the predicted trajectory as $\trajhat$.

\gapheading{Missing trajectory-level, goal-conditioned supervision.}
\label{gap:trajectory-supervision}
The prediction loss constrains local transitions, whereas SIGReg shapes the distribution of latent representations. Neither term in \Cref{eq:ldyn} jointly uses an ordered complete trajectory and its task goal. The original objective therefore provides no explicit trajectory-level, goal-conditioned supervision to the shared representation.

\gapheading{Endpoint scoring is insensitive to intermediate paths.}
\label{gap:endpoint-scoring}
\Cref{eq:c0} reads only the predicted endpoint. For any two intermediate latent-state sequences $u_{1:H-1}$ and $v_{1:H-1}$ that share an initial state $z_0$ and endpoint $z_H$,
\begin{equation}
\label{eq:endpoint-invariance}
\Czero\big((z_0,u_{1:H-1},z_H),g\big)
=\Czero\big((z_0,v_{1:H-1},z_H),g\big).
\end{equation}
The endpoint term therefore cannot directly use information about how a candidate reaches its endpoint. Intermediate dynamics may nevertheless provide additional evidence about candidate quality under the specified goal when endpoint distance alone is insufficient. Together, these gaps leave trajectory-level, goal-conditioned information without an explicit role in either representation learning or candidate scoring.

\subsection{Goal-Conditioned Trajectory Cost}
\label{sec:method:functional}

\paragraph{Designing a goal-conditioned path functional.}
To use information along the complete predicted path rather than only
its endpoint, we formulate trajectory quality as a goal-conditioned path
functional over an ordered sequence of latent states and instantiate it
as a latent trajectory cost (LTC). LTC maps a predicted trajectory to a learned scalar cost by
aggregating latent states, their temporal changes, and goal-relative
information over time. Intermediate-path information therefore
contributes directly to trajectory evaluation rather than being
discarded by endpoint-only scoring. The ranking semantics of this cost
are learned from trajectory preferences: goal-matched expert
trajectories should receive lower costs than synthetic negative
trajectories and execution failures.

We implement this functional using four components:
\textbf{(i) latent state and local evolution}, represented by the current
state $\latent_t$ and the first difference
$\Delta\latent_t=\latent_{t+1}-\latent_t$ between adjacent latent states;
\textbf{(ii) goal conditioning}, introduced at every step through the
goal-relative quantity $d_t^g=\goal-\latent_t$;
\textbf{(iii) temporal position}, represented by the normalized phase
$\eta_t=t/\max(T-1,1)$; and
\textbf{(iv) learned trajectory aggregation}, in which $\ell_\psi$ maps
the information at each step to a nonnegative contribution and these
contributions are averaged over the trajectory. The goal-relative term
allows the same latent state and change to receive different evaluations
under different goals, while the phase term distinguishes where an
evolution occurs along the trajectory. Averaging over time produces a
length-normalized cost for the complete trajectory.

\begin{definition}[Goal-Conditioned Latent Trajectory Cost]
Given a latent trajectory $\traj=(\latent_0,\ldots,\latent_T)$ with
$T\ge1$ and goal representation $\goal$, define
\begin{equation}
\label{eq:spsi}
\Spsi(\traj\mid g)
=\frac{1}{T}\sum_{t=0}^{T-1}\ell_\psi\!\left(
\latent_t,\Delta\latent_t,d_t^g,\eta_t
\right).
\end{equation}
Here $\ell_\psi:\R^{3d+1}\!\to\!\R_{\ge0}$ is a learnable nonnegative
per-step contribution function, whose aggregation over time defines the
cost of the complete trajectory.
\end{definition}

Under the learned trajectory preferences, a lower $\Spsi$ indicates a
trajectory preferred as goal-matched expert behavior, whereas a higher
value indicates a less preferred trajectory, such as a synthetic
negative or an execution failure. Unlike $\Czero$, which reads only the
endpoint, LTC is structurally able to use intermediate latent states,
their changes, and their relations to the goal.

\subsection{Trajectory Preference Supervision}
\label{sec:method:pref}

Write a goal-conditioned example as $x=(\traj,g)$ and abbreviate $\Spsi(x)=\Spsi(\traj\mid g)$. For a preference pair $(x^+,x^-)$, define the cost margin as $m_\psi(x^+,x^-)=\Spsi(x^-)-\Spsi(x^+)$ and use the pairwise logistic loss
\begin{equation}
\label{eq:pairwise}
\ell_\beta(x^+,x^-)
=\log\!\left(1+\exp\!\left[-\frac{m_\psi(x^+,x^-)}{\beta}\right]\right),
\end{equation}
where $\beta>0$ is a temperature. Minimizing \Cref{eq:pairwise} encourages $\Spsi(x^+)<\Spsi(x^-)$, so that a goal-matched expert trajectory is preferred to its corresponding negative.

\paragraph{Synthetic preferences.}
\label{sec:method:negatives}
For an expert trajectory $\trajp=(\latent_0^+,\ldots,\latent_T^+)$, let the goal representation be $\goal=\latent_T^+$ and define the positive example $x^+=(\trajp,g)$. The first negative replaces the goal with a batchwise cyclically shifted goal $g'$:
\begin{equation}
\label{eq:gm}
x^-_{\mathrm{gm}}=(\trajp,g').
\end{equation}
The second negative adds noise to the intermediate states. Draw $\epsilon_t\sim\mathcal N(0,\sigma^2s_{\mathrm{batch}}^2I)$ and fix $\epsilon_0=\epsilon_T=0$; the perturbed states are
\begin{equation}
\label{eq:jit}
\latent_t^-=\latent_t^++\epsilon_t,\qquad 0\le t\le T.
\end{equation}
Let $\trajn_{\mathrm{jit}}=(\latent_0^-,\ldots,\latent_T^-)$ and $x^-_{\mathrm{jit}}=(\trajn_{\mathrm{jit}},g)$. Goal mismatch trains LTC to assign an expert trajectory a lower cost with its corresponding goal than with an unrelated goal. Endpoint-preserving perturbations train LTC to distinguish different intermediate paths even when the initial and final states are fixed.

We stop gradients through the synthetic negative branch while retaining the current encoder's computation graph for the expert branch. Synthetic preferences therefore update LTC directly and update the encoder through the expert branch; their effect on negative encodings is mediated indirectly by the cost landscape learned by LTC. The corresponding loss is
\begin{equation}
\label{eq:lpath}
\Lpath
=\E\!\left[
\frac{
w_{\mathrm{gm}}\ell_\beta(x^+,x^-_{\mathrm{gm}})
+w_{\mathrm{jit}}\ell_\beta(x^+,x^-_{\mathrm{jit}})}
{w_{\mathrm{gm}}+w_{\mathrm{jit}}}
\right].
\end{equation}

\paragraph{Closed-loop failure preferences.}
Synthetic negatives expose predefined structural differences but cannot
cover errors produced by the model during actual planning. After each
training epoch, we therefore use the current model to perform
endpoint-only CEM planning and execute the resulting plan in a resettable
training environment. We deliberately omit LTC during failure mining so
that it does not filter out failures selected by endpoint-only scoring;
these episodes provide negative examples of planning errors that the
endpoint term alone cannot identify. The endpoint-only planner is used
only for collecting training failures, whereas final planning uses the
joint endpoint-plus-LTC score defined below. The environment's success
criterion supplies a binary episode-level label. For each failed episode
$i$, we encode the executed observations frame by frame as
$\trajn_i=(\enc(o^{\mathrm{exec}}_{i,0}),\ldots,
\enc(o^{\mathrm{exec}}_{i,T_i^-}))$ and pair it with an expert trajectory
$\trajp_i$ having the same initial condition and goal:
\begin{equation}
\label{eq:mine}
(x_i^+,x_i^-)
=\big((\trajp_i,g_i),(\trajn_i,g_i)\big).
\end{equation}
Positive and negative trajectories may differ in length because
\Cref{eq:spsi} averages each trajectory over its own number of
transitions. We insert these preference pairs sequentially into a
bounded FIFO buffer $\mathcal B$. The corresponding buffer loss is
\begin{equation}
\label{eq:lmined}
\Lmined
=\E_{(x^+,x^-)\sim\mathcal B}
\ell_\beta(x^+,x^-).
\end{equation}
The trajectory and goal representations inserted into the buffer are
detached, so $\Lmined$ updates only LTC in the current backward pass. It
first changes LTC parameters and then indirectly influences the encoder
through $\Lpath$ in subsequent batches.

We intentionally exclude the mined observations from predictor training,
because using them as transition targets would introduce online dynamics
adaptation and confound it with the effect of LTC. Closed-loop mining is
instead an integral part of the proposed trajectory-preference module:
it supplies failure preferences only for calibrating LTC, while LeWM's
predictor data and objective remain unchanged.
\subsection{Joint Training and Gradient Flow}
\label{sec:method:objective}

Writing the LeWM objective defined in \Cref{eq:ldyn} as $\mathcal L_{\mathrm{LeWM}}$, the complete training objective is
\begin{equation}
\label{eq:total}
\Lvalewm
=\mathcal L_{\mathrm{LeWM}}
+\lambda_{\mathrm{path}}\Lpath
+\lambda_{\mathrm{mined}}\Lmined.
\end{equation}
\Cref{tab:grad} summarizes the direct gradient flow from each loss in one backward pass. This design preserves the predictor's training signal while allowing trajectory preferences to shape the shared encoder through the synthetic expert branch. Closed-loop failure preferences primarily calibrate LTC against the model's own failure modes, while the synthetic expert branch supplies trajectory-level supervision to the encoder.

\begin{table}[t]
\centering\small
\begin{tabular}{lccc}
\toprule
Loss & Encoder $\enc$ & Predictor $\pred$ & LTC $\Spsi$\\
\midrule
$\Lpred$ & \checkmark & \checkmark & \\
$\Lsig$ & \checkmark & & \\
$\Lpath$ & \checkmark & & \checkmark\\
$\Lmined$ & & & \checkmark\\
\bottomrule
\end{tabular}
\caption{Direct gradient flow from each loss term.}
\label{tab:grad}
\end{table}

Each epoch has two stages. First, we jointly optimize \Cref{eq:total} on offline expert batches: we construct goal-mismatched and endpoint-preserving perturbations to compute $\Lpath$ and, when the buffer is nonempty, sample closed-loop preferences to compute $\Lmined$. We then disable gradients, perform closed-loop planning with the current model, and add newly discovered failed trajectories to $\mathcal B$. Training alternates between these two stages.

\subsection{Trajectory-Sensitive Candidate Scoring}
\label{sec:method:margin}

During planning, LTC reads, as defined in \Cref{eq:context-rollout}, the context-augmented predicted latent trajectory $\trajhat$. To eliminate the scale difference between LTC output and endpoint distance, we first define the calibrated LTC score
\begin{equation}
\label{eq:ltc-scale}
\Stilde(\trajhat\mid g)
=
\frac{\operatorname{IQR}(\Czero)}
{\max\!\left(\operatorname{IQR}(\Spsi),\epsilon\right)}
\Spsi(\trajhat\mid g).
\end{equation}
Here $\epsilon>0$ is a small constant for numerical stability. $\operatorname{IQR}(x)=Q_{0.75}(x)-Q_{0.25}(x)$ denotes the interquartile range; the interquartile ranges of $\Czero$ and $\Spsi$ are both estimated over the same set of candidates produced by endpoint-only CEM. The combined score is defined as
\begin{equation}
\label{eq:combined}
\Clambda(\trajhat,g)
=\Czero(\trajhat,g)
+\lambda\Stilde(\trajhat\mid g),
\end{equation}
where $\Czero$ evaluates only the predicted endpoint, $\Stilde$ reads the complete predicted latent trajectory, and $\lambda\ge0$ controls the relative strength of the LTC signal. When $\lambda=0$, candidates are ranked solely by the endpoint term, although the encoder may still have been shaped by trajectory preference training.

Thus, the combined score retains endpoint-based goal matching while incorporating information from complete predicted trajectories into candidate selection.
 \section{Experiments}
\label{sec:experiments}

We first evaluate the closed-loop planning performance of \method across four simulated environments. We then organize controlled experiments around the two gaps identified in the motivation: Gap~1 tests how trajectory-preference supervision affects the shared representation, while Gap~2 tests whether LTC uses intermediate paths and whether this information improves candidate ranking. We further report a real-robot evaluation. Finally, ablations separate the effects of trajectory supervision during training and explicit LTC-based ranking during planning.

\subsection{Main Results: Closed-Loop Planning in Simulation}
\label{sec:exp:main}

\paragraph{Experimental setup.}
We evaluate closed-loop planning using the public LeWM datasets~\citep{maes_lelidec2026lewm}
for Push-T~\citep{chi2025diffusionpolicy,zhou2024dinowm},
OGBench-Cube~\citep{park2024ogbench},
Reacher~\citep{tassa2018dmcontrol}, and
Two-Room~\citep{sobal2025learning}, covering object manipulation,
continuous control, and obstacle navigation.
LeWM~\citep{maes_lelidec2026lewm} is our primary
controlled baseline. LeWM and \method use the same offline dataset,
encoder--predictor backbone, predictor objective, optimization schedule,
MPC framework, and CEM sampling budget. Their intended planning-time
difference is the candidate-scoring objective: LeWM uses endpoint distance,
whereas \method combines endpoint distance with LTC.

\method additionally collects a small number of closed-loop failures solely
to construct preference pairs for LTC; these trajectories are not added to
the predictor's transition-training data. Accordingly, the comparison holds
the predictor's transition data and learning objective fixed and evaluates
the effect of introducing the proposed trajectory-preference module into the
LeWM framework.

Following LeWM, we report task success rates under a shared evaluation
protocol. For broader comparison, \Cref{tab:main-results} also includes
PLDM~\citep{sobal2025learning}, DINO-WM~\citep{zhou2024dinowm}, and the
goal-conditioned methods GCBC, GCIVL, and
GCIQL~\citep{park2024ogbench,kostrikov2022iql}. Additional implementation and
evaluation details, including the model architecture, training, failure
mining, planning settings, and multi-seed evaluation counts, are provided in
the appendix.

\paragraph{Main results.}
With joint endpoint-plus-LTC scoring on every task, \method achieves mean success rates of $99\%$, $88\%$, $93\%$, and $94\%$ on Push-T, Cube, Reacher, and Two-Room, respectively, averaged over three evaluation seeds. Compared with LeWM, these results correspond to gains of $3$, $14$, $7$, and $7$ percentage points. Thus, \method improves closed-loop planning across object manipulation, continuous control, and navigation tasks.
\begin{table}[t]
\centering
\small
\setlength{\tabcolsep}{3.2pt}
\begin{tabular}{lcccc}
\toprule
Model & Push-T & Cube & Reacher & Two-Room\\
\midrule
GCBC    & 75 & 84 & -- & \textbf{100}\\
GCIVL   & 33 & 56 & -- & \textbf{100}\\
GCIQL   & 20 & 64 & -- & \textbf{100}\\
PLDM    & 78 & 65 & 78 & 97\\
DINO-WM & 74 & 86 & 79 & \textbf{100}\\
LeWM    & 96 & 74 & 86 & 87\\
\rowcolor[gray]{0.94}
\method~(Ours) & \textbf{99} & \textbf{88} & \textbf{93} & 94\\
\bottomrule
\end{tabular}
\caption{Closed-loop planning success rates (\%) across four environments.
For \method, results are averaged over three evaluation seeds
 and rounded to the nearest percentage point.
Bold indicates the best result for each task.}
\label{tab:main-results}
\end{table}

\paragraph{Analysis.}
\method outperforms LeWM on all four tasks and achieves the best results in the table on Push-T, Cube, and Reacher. It uses an end-to-end trained lightweight encoder, and LTC adds fewer than $1$M parameters, leaving the overall model in the same parameter regime as LeWM. These results show that goal-conditioned trajectory preferences and path-sensitive evaluation improve planning within a lightweight world model.

\subsection{Controlled Tests of the Two Gaps}
\label{sec:exp:gap-validation}

We organize the following controlled experiments around the two gaps identified in the motivation. For \Cref{gap:trajectory-supervision}, we exclude LTC from planning and compare endpoint-only rankings between LeWM and \method, isolating the effect of trajectory-preference supervision on the shared representation. For \Cref{gap:endpoint-scoring}, we first test whether LTC improves candidate ranking when endpoint predictions are inaccurate, and then control endpoint information to verify that LTC uses intermediate paths. Execution outcomes are obtained by running the corresponding candidate action sequences in the environment.

\begin{figure}[t]
\centering
\includegraphics[width=\linewidth]{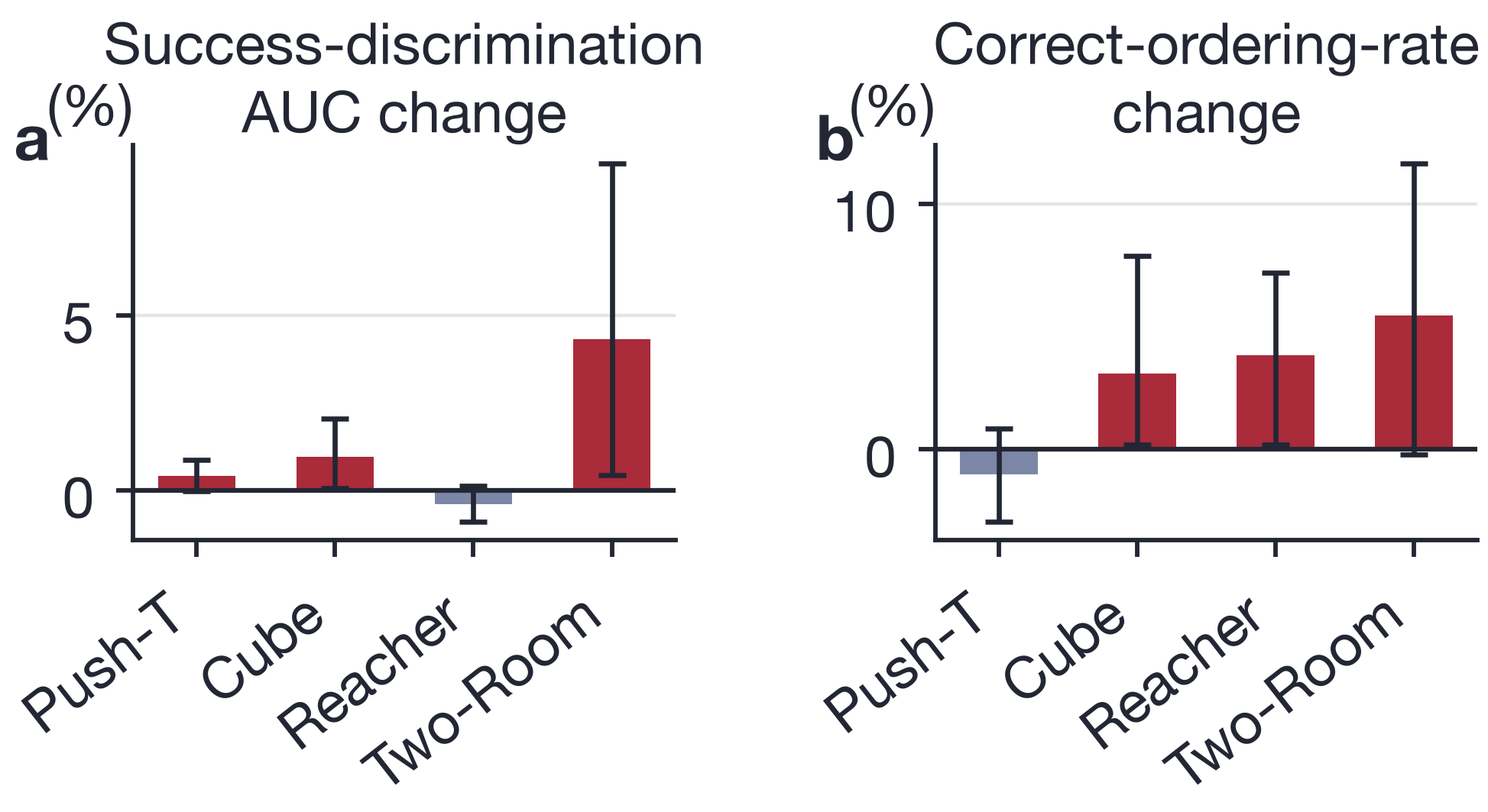}
\caption{Controlled ranking tests. \textbf{(a)} Endpoint-only AUC change from LeWM to \method. \textbf{(b)} Correct-ordering change after adding LTC for candidates with large prediction--execution discrepancies.}
\label{fig:decision-alignment-comparison}
\end{figure}

\paragraph{Testing \Cref{gap:trajectory-supervision}: representation shaping.}
To test \Cref{gap:trajectory-supervision}, we conduct a controlled experiment using the same candidate action sequences for both models. We first execute each sequence in the environment to obtain its ground-truth success or failure label. We then remove LTC from the planning score and use LeWM and \method to score the candidates solely by predicted endpoint distance in their respective representation spaces. By comparing these endpoint scores with the execution labels, we compute the success-discrimination AUC for each model. Because the candidate sequences and scoring form are held fixed, the AUC difference reflects the effect of trajectory-preference supervision on the shared representation.

Compared with LeWM, \method improves the endpoint-score AUC by $0.4$, $1.0$, and $4.3$ percentage points on Push-T, Cube, and Two-Room, respectively, with statistically significant gains on Cube and Two-Room. The change on Reacher is a nonsignificant $-0.4$ percentage points (\Cref{fig:decision-alignment-comparison}a). These results show that trajectory-preference supervision can shape the shared encoder and improve the ability of endpoint scores to distinguish successful from failed executions.

\paragraph{Testing \Cref{gap:endpoint-scoring}: path use and candidate ranking.}
We evaluate LTC from two perspectives. First, we test whether LTC improves candidate ranking when endpoint predictions are inaccurate. We sample a set of candidate action sequences and use \method to roll out each sequence in latent space. For each candidate, we measure the latent-space discrepancy between its predicted endpoint and the endpoint obtained by executing the same action sequence in the environment and encoding the resulting observation. Within each start--goal instance, we construct success--failure candidate pairs and focus on pairs with large prediction--execution discrepancies. We then compare endpoint-only scoring with joint endpoint-plus-LTC scoring and measure how often the successful candidate is ranked ahead of the failed candidate. Adding LTC improves the correct-ordering rate by $3.1$, $3.8$, and $5.5$ percentage points on Cube, Reacher, and Two-Room, respectively, while the change on Push-T is $-1.0$ percentage points (\Cref{fig:decision-alignment-comparison}b). These results show that when endpoint predictions are inaccurate, LTC improves endpoint-based ranking on three of the four tasks.

\begin{figure}[t]
\centering
\includegraphics[width=\linewidth]{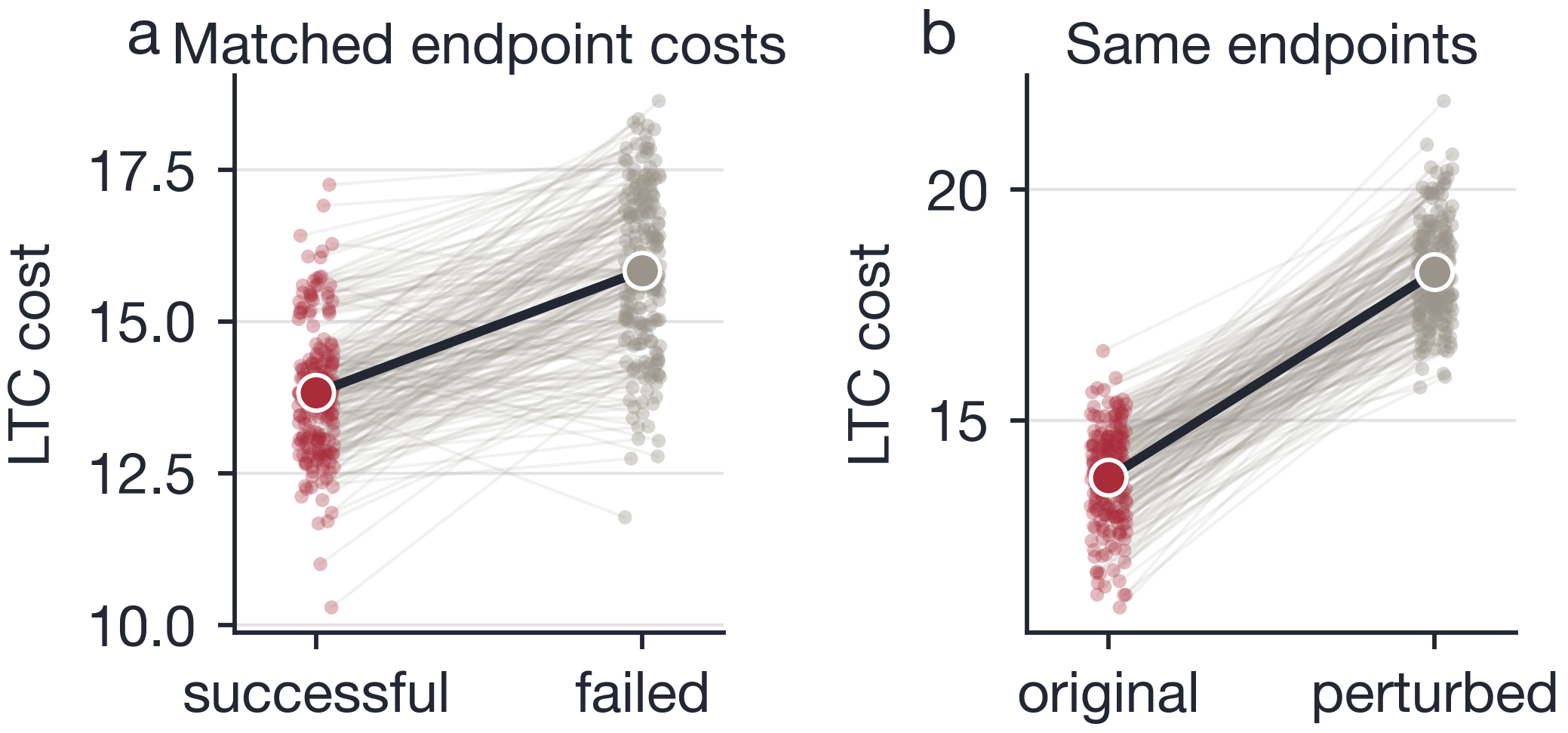}
\caption{Endpoint-controlled tests on Reacher. \textbf{(a)} LTC correctly ranks $94.4\%$ of $207$ endpoint-matched success--failure pairs. \textbf{(b)} Intermediate-state perturbations increase LTC cost in all $200$ pairs with fixed endpoints.}
\label{fig:reachability-main}
\end{figure}

We next control for endpoint information to directly test whether LTC
uses intermediate paths. Within each start--goal instance, we match
successful and failed executions one-to-one according to their \method
endpoint costs. We retain only the hardest pairs, whose endpoint-cost
difference is at most $0.25$ times the within-instance interquartile
range. On Reacher, we evaluate $1{,}344$ candidates, obtain $395$
matched success--failure pairs, and retain the $207$ hardest pairs under
this criterion. When start--goal instances are weighted equally, LTC
ranks the successful trajectory ahead of the failed trajectory with
correct-ordering rates of $85.8\%$, $90.0\%$, $94.4\%$, and $83.0\%$
on Push-T, Cube, Reacher, and Two-Room, respectively;
\Cref{fig:reachability-main}a shows Reacher as an example.
We then fix the initial and final states of each trajectory and perturb
only its intermediate latent states, giving the original and perturbed
trajectories identical endpoint costs. Across $200$ comparisons per
task, LTC assigns a higher cost to the perturbed trajectory in $100\%$
of the comparisons on Push-T, Cube, and Reacher, and in $96\%$ on
Two-Room; \Cref{fig:reachability-main}b shows Reacher as an example.
Together, these endpoint-controlled tests demonstrate that LTC uses
intermediate-path information unavailable to endpoint-only scoring.

\subsection{Real-Robot Evaluation}
\label{sec:real-robot}

We evaluate goal-conditioned visual planning on a physical Franka FR3
using real manipulation trajectories recorded at $30$\,Hz from three
synchronized RGB views. The evaluation contains 20 short-horizon start--goal tasks, each evaluated by both methods. Each task asks the robot to move from a recorded
initial observation toward the visual goal observed $1.5$ seconds later
in the same trajectory; the corresponding seven-joint configuration is
used to measure goal-reaching accuracy. LeWM and \method receive the same
start and goal in every task. A rollout is considered successful if it
avoids collision and its minimum seven-joint $L_2$ distance to the goal
configuration does not exceed $0.11\,\mathrm{rad}$. A completed rollout
above this threshold is counted as a target miss. 

\begin{figure}[ht]
    \centering
    \includegraphics[width=\columnwidth]{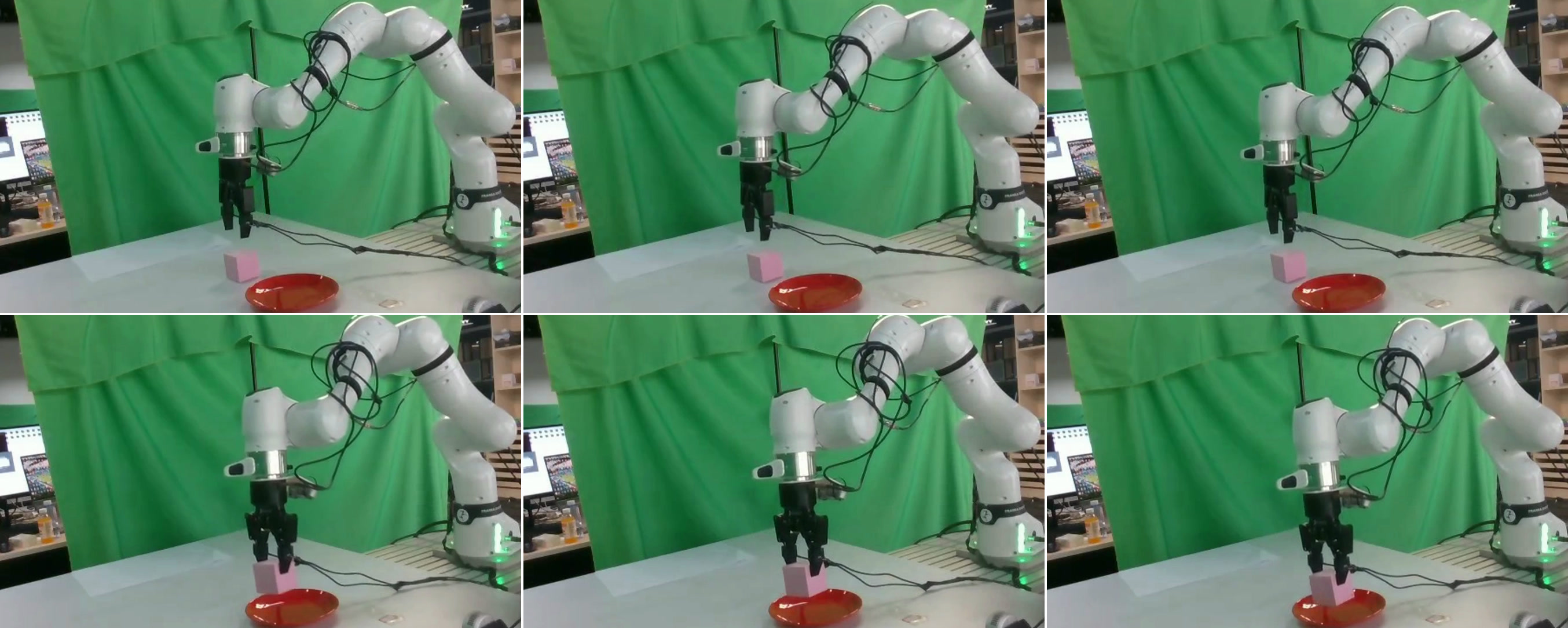}
    \caption{Real-robot setup and two example start--goal tasks. Each row shows the initial observation, an intermediate state, and the visual goal $1.5$ seconds later.}
    \label{fig:real-robot-tasks}
\end{figure}

\begin{table}[ht]
    \centering
    \small
    \setlength{\tabcolsep}{4.0pt}
    \begin{tabular}{@{}lccc@{}}
        \toprule
        Model
        & Success $\uparrow$
        & Target miss $\downarrow$
        & Collision $\downarrow$ \\
        \midrule
        LeWM
        & 10/20 (50\%)
        & 6/20 (30\%)
        & 4/20 (20\%) \\
        Traj-LeWM
        & \textbf{14/20 (70\%)}
        & \textbf{4/20 (20\%)}
        & \textbf{2/20 (10\%)} \\
        \bottomrule
    \end{tabular}
    \caption{Real-robot outcomes for LeWM and Traj-LeWM on the same 20 start--goal tasks using a Franka FR3.}
    \label{tab:real-robot}
\end{table}

As shown in \Cref{tab:real-robot}, \method succeeds in 14 of the
20 trials, compared with 10 for LeWM. Target misses decrease from six
with LeWM to four with \method, while collision-induced terminations
decrease from four to two. Given the limited number of paired tasks, we interpret these outcomes as preliminary evidence of real-robot feasibility rather than a conclusive performance advantage.

\subsection{Ablation Studies}
\label{sec:exp:ablation}

\Cref{tab:path-cost} disentangles the method along two dimensions. The first three rows use the same endpoint score and vary the training supervision, isolating how trajectory preferences affect the shared latent model. The last three rows use the same trained \method model and vary the planning score, isolating the contribution of explicit LTC-based ranking.

\begin{table}[ht]
    \centering
    \small
    \setlength{\tabcolsep}{1.2pt}
    \begin{tabular}{@{}llccccc@{}}
    \toprule
    Training & Score & Push-T & Cube & Reacher & Two-Room & Avg.\\
    \midrule
    LeWM       & Endpoint & 96 & 74 & 86 & 87 & 85.8\\
    \methodsyn & Endpoint & 96 & 79 & 80 & \textbf{94} & 87.3\\
    \method    & Endpoint & 98 & 80 & 84 & \textbf{94} & 89.0\\
    \method    & LTC   & 78 & 69 & \textbf{94} & 80 & 80.3\\
    \method    & Joint & \textbf{99} & \textbf{88} & 93 & \textbf{94} & \textbf{93.5}\\
    \bottomrule
    \end{tabular}
    \caption{Ablation of training supervision and planning score (success rate, \%). Bold indicates the best result for each task.}
    \label{tab:path-cost}
\end{table}

\paragraph{Training-time supervision.}
Under endpoint-only scoring, synthetic preferences raise the average success rate from $85.8\%$ to $87.3\%$, and adding closed-loop failures raises it to $89.0\%$. Because LTC is absent at test time, these gains reflect trajectory supervision of the shared representation.

\begin{figure}[ht]
    \centering
    \includegraphics[width=\linewidth]{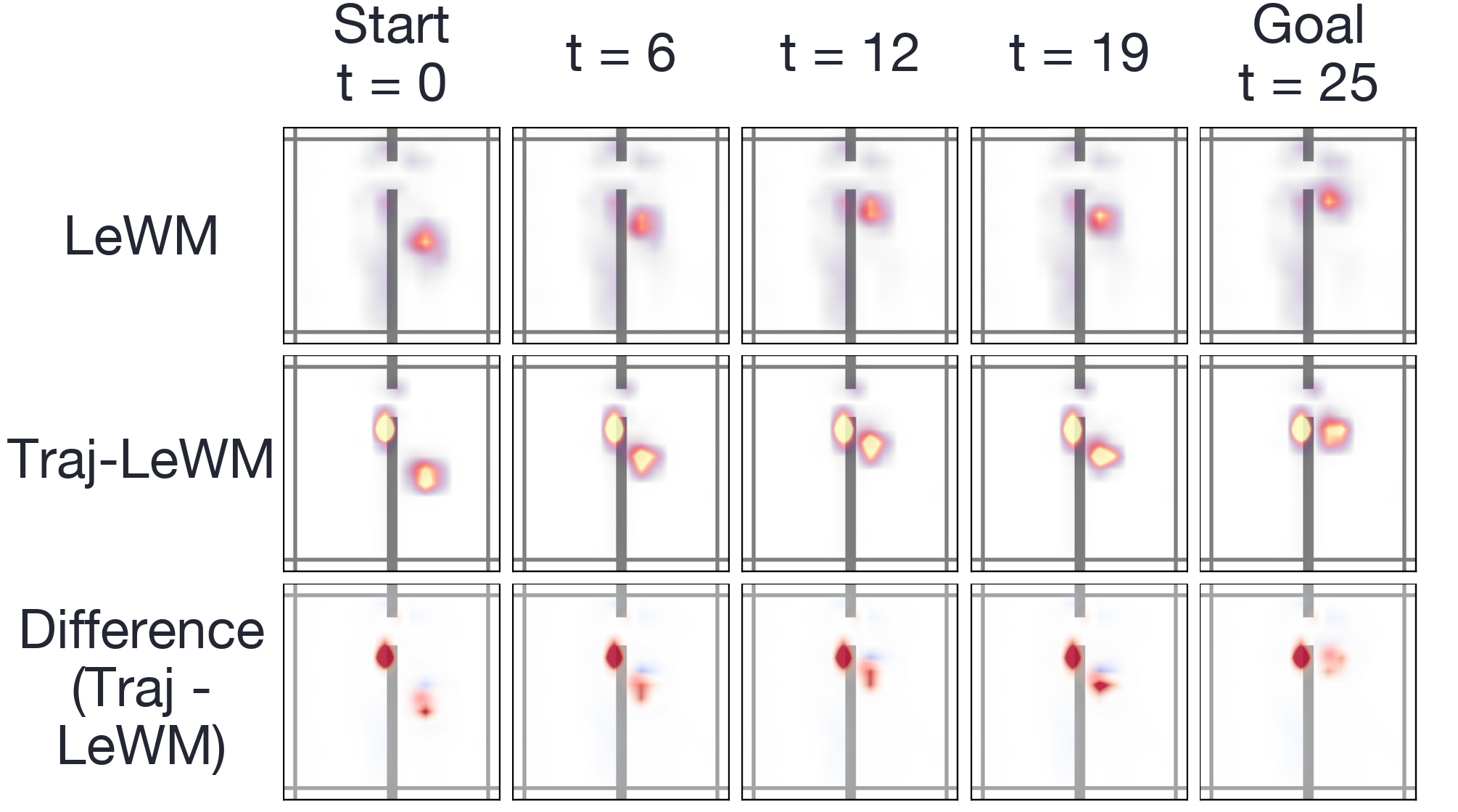}
    \caption{Final-layer CLS-to-patch attention along a Two-Room trajectory. Red indicates higher attention in Traj-LeWM/LeWM.}
    \label{fig:two_room_attention}
\end{figure}

The attention comparison provides a qualitative view of this representation effect. Along the same Two-Room trajectory, LeWM's final-layer CLS-to-patch attention largely follows the agent, whereas \method additionally responds to the central partition and doorway boundary (\Cref{fig:two_room_attention}). This redistribution is consistent with the quantitative endpoint-only results: trajectory supervision changes the spatial evidence emphasized by the shared encoder even when LTC is absent from the planning score.

\paragraph{Effect of planning-time LTC scoring.}
With the trained \method model fixed, LTC-only ranking achieves an average success rate of $80.3\%$ and reaches $94\%$ on Reacher. Thus, the learned path score can independently provide a useful planning signal. Combining LTC with the endpoint term raises the average from $89.0\%$ under endpoint-only scoring to $93.5\%$, including gains from $80\%$ to $88\%$ on Cube and from $84\%$ to $93\%$ on Reacher.  Across tasks, the ablation shows that the two components contribute
differently. Trajectory supervision mainly improves Cube and Two-Room
under endpoint scoring, whereas LTC-only scoring is strongest on
Reacher. This task-dependent behavior further suggests that intermediate-path evidence is most valuable when endpoint distance alone provides an incomplete ranking signal. The joint score achieves the highest average success rate,
indicating that endpoint distance and LTC are complementary rather than
interchangeable ranking signals.
\section{Conclusion}
\label{sec:conclusion}

This work addresses two limitations of goal-conditioned world models: local prediction objectives lack trajectory-level constraints, and endpoint distance ignores intermediate paths. We propose \method, which learns a goal-conditioned Latent Trajectory Cost (LTC) from trajectory-level preferences and uses trajectory information for both representation learning through the shared encoder and candidate ranking during planning. With joint endpoint-plus-LTC scoring, \method improves success rates over LeWM by $14$, $7$, $7$, and $3$ percentage points on Cube, Two-Room, Reacher, and Push-T, respectively. Controlled tests show that trajectory-preference supervision improves the ability of endpoint scores to discriminate execution outcomes on three tasks. They also show that LTC distinguishes trajectories when endpoint information is similar or identical and improves the ranking of candidates with inaccurate endpoint predictions on Cube, Reacher, and Two-Room. Ablations further verify the complementary roles of representation shaping and explicit path-sensitive ranking.

\crefname{equation}{Eq.}{Eqs.}
\Crefname{equation}{Eq.}{Eqs.}
\crefname{figure}{Fig.}{Figs.}
\Crefname{figure}{Fig.}{Figs.}

\providecommand{\method}{Traj-LeWM\xspace}
\providecommand{\methodsyn}{Traj-LeWM-syn\xspace}

\renewcommand{\thesection}{\Alph{section}}
\renewcommand{\thesubsection}{\Alph{section}.\arabic{subsection}}
\renewcommand{\thefigure}{\arabic{figure}}
\renewcommand{\thetable}{\arabic{table}}
\renewcommand{\theequation}{\arabic{equation}}

\setcounter{secnumdepth}{2}

\clearpage
\onecolumn
\setcounter{section}{0}
\setcounter{subsection}{0}
\begin{center}
{\Large\bfseries Appendix}
\end{center}
\medskip

\section{Reproducibility Details}
\label{app:reproducibility}

\subsection{Model, Training, and Mining}
\label{app:model-training}

\paragraph{Architecture.}
The visual encoder is a ViT with $14\times14$ image patches, hidden dimension
192, 12 Transformer blocks, 3 attention heads, and MLP dimension 768. The
autoregressive predictor contains 6 Transformer blocks, 16 attention heads,
head dimension 64, MLP dimension 2048, and dropout 0.1. The latent trajectory
cost (LTC) takes
\[
  [z_t,\;z_{t+1}-z_t,\;z_g-z_t,\;t/(T-1)]
\]
as input and applies three width-512 hidden layers, each with LayerNorm, GELU,
and dropout 0.1, followed by a positive scalar output and temporal averaging.
For the retained configurations with a 10-dimensional model action, LeWM has
18,042,672 parameters, while Traj-LeWM-syn and Traj-LeWM each have 18,867,505
parameters; the LTC contributes 824,833 additional parameters. Cube's
25-dimensional model action adds 2,880 action-embedding parameters to each
total.

\paragraph{Optimization.}
We use AdamW with learning rate $5\times10^{-5}$, weight decay $10^{-3}$,
batch size 128, bf16 mixed-precision training, gradient clipping at 1.0, and a
linear-warmup cosine-annealing scheduler. The trajectory-supervision
configuration uses $\lambda_{\mathrm{path}}=\lambda_{\mathrm{mined}}=0.05$,
logistic-loss temperature $\beta=0.2$, SIGReg weight 0.09, 17 knots, and
1,024 projections. The synthetic goal-mismatch and jitter terms have weights
1.0 and 0.5, respectively; the jitter scale is 0.05 times the scalar latent
standard deviation computed over the current batch and time dimensions. 

\paragraph{Failure mining.}
After each training epoch, the current model evaluates 200 start--goal
queries using endpoint-only CEM.  If a rollout does not reach the environment termination
condition within 50 steps, it provides a negative trajectory; the
corresponding dataset segment with the same start and goal provides the
positive trajectory. Detached  latent pairs are stored in a FIFO buffer
with capacity 2,048.

\subsection{Planning}
\label{app:planning}

Formal evaluation uses MPC with CEM. Each planning round samples 300 action
sequences, retains 30 elites, and updates the Gaussian mean and standard
deviation from those elites. 
The initial CEM distribution has zero mean and unit variance in standardized
action coordinates. Subsequent means and standard deviations are the
empirical elite statistics, with warm starts enabled across replanning
rounds. 

For joint planning, the retained implementation uses
\[
\begin{aligned}
 C_{\mathrm{joint}}&=C_{\mathrm{endpoint}}+
 w_{\mathrm{LTC}} C_{\mathrm{LTC}},\\
 w_{\mathrm{LTC}}&=\lambda
 \frac{\operatorname{IQR}(C_{\mathrm{endpoint}})}
 {\max\{\operatorname{IQR}(C_{\mathrm{LTC}}),10^{-12}\}}.
\end{aligned}
\]
The IQRs are estimated from endpoint-only calibration candidates. The
final joint-scoring configuration adopted for the paper is
$\lambda=(0.5,0.8,0.6,0.5)$ for Push-T, Cube, Reacher, and Two-Room,
respectively. 

\section{Additional Results and Diagnostics}
\label{app:additional-results}

\subsection{Verified Multi-Seed Evaluation}
\label{app:verified-main}

Table~\ref{tab:n50} reports the  Traj-LeWM results for evaluation
seeds 42--44. The success-count column lists the number of successes out of
50 queries for each seed. Success rates are the mean and sample standard
deviation across the three evaluation seeds. 

\begin{center}
\centering
\small
\setlength{\tabcolsep}{3pt}
\begin{tabular}{lcc}
\toprule
Task & Successes (42/43/44) & Success rate (mean $\pm$ SD) \\
\midrule
Push-T   & 49/50/49 & $98.7\pm1.2$ \\
Cube     & 44/46/42 & $88.0\pm4.0$ \\
Reacher  & 46/47/46 & $92.7\pm1.2$ \\
Two-Room & 47/45/49 & $94.0\pm4.0$ \\
\bottomrule
\end{tabular}
\captionof{table}{Epoch-10 Traj-LeWM results on 50 start--goal queries per
task and evaluation seed. Success rates are mean $\pm$ sample standard
deviation across evaluation seeds 42--44, in percent.}
\label{tab:n50}
\end{center}

\subsection{Controlled Path-Ranking Diagnostics}
\label{app:path-diagnostics}

\paragraph{Metric definitions.}
\textbf{Success rate.}
Let $s_q=1$ if query $q$ terminates successfully and $s_q=0$ otherwise. For
$N$ queries, the success rate is the percentage of successful queries:
\[
  \frac{100}{N}\sum_{q=1}^{N}s_q.
\]

\textbf{Endpoint ROC--AUC.}
Some diagnostic queries contain multiple successful and failed candidate
trajectories. For each such query, candidates are scored by
$-C_{\mathrm{endpoint}}$, so a lower endpoint cost gives a higher score. The
query-level ROC--AUC measures how often a successful candidate is ranked above
a failed candidate; a tie contributes one half. A query containing only
successful candidates or only failed candidates is excluded because its
ROC--AUC is undefined. We report the arithmetic mean over the remaining
queries, giving every query equal weight.

\textbf{LTC pair-ordering accuracy.}
The endpoint-matched analysis forms pairs containing one successful and one
failed executed trajectory with similar endpoint costs. A pair receives a
score of one if LTC assigns lower cost to the successful trajectory, zero if
it assigns lower cost to the failed trajectory, and one half for a tie. The
reported percentage is 100 times the mean score over all retained pairs.

\paragraph{Endpoint-matched executed trajectories.}
For each start--goal query, successful executed trajectories are paired
one-to-one with unused failed trajectories having the nearest endpoint
energy, subject to a normalized endpoint-gap caliper of 0.25. Table
\ref{tab:endpoint-matched} reports the percentage of pairs for which LTC
assigns lower cost to the successful trajectory. We omit confidence intervals
because the corrected aggregate count is insufficient to
reconstruct the previous query-level bootstrap interval.

\begin{center}
\centering
\small
\begin{tabular}{lrrc}
\toprule
Task & Queries & Pairs & Correct ordering (\%) \\
\midrule
Push-T   & 21 &  94 & 85.8 \\
Cube     & 15 &  30 & 90.0 \\
Reacher  & 22 & 207 & 94.4 \\
Two-Room & 18 & 106 & 83.0 \\
\bottomrule
\end{tabular}
\captionof{table}{LTC ordering accuracy, in percent, for endpoint-matched
successful and failed executed trajectories.}
\label{tab:endpoint-matched}
\end{center}

\subsection{Candidate-Selector Agreement}
\label{app:selector-overlap}

We test whether LTC merely reproduces endpoint ranking by applying three
selectors to the same candidate action sequences: LeWM endpoint distance,
Traj-LeWM endpoint distance, and LTC alone. For selectors $a$ and $b$, exact
top-1 agreement indicates whether they choose the same minimum-cost candidate,
while top-5 intersection is
$|S_a^{(5)}\cap S_b^{(5)}|/5$. Metrics are averaged equally across queries,
and 95\% confidence intervals use 20,000 bootstrap samples of complete queries.
Queries for this analysis are drawn from evaluation seeds 42--44 across all tasks.

\begin{figure*}[t]
\centering
\includegraphics[width=0.96\textwidth]{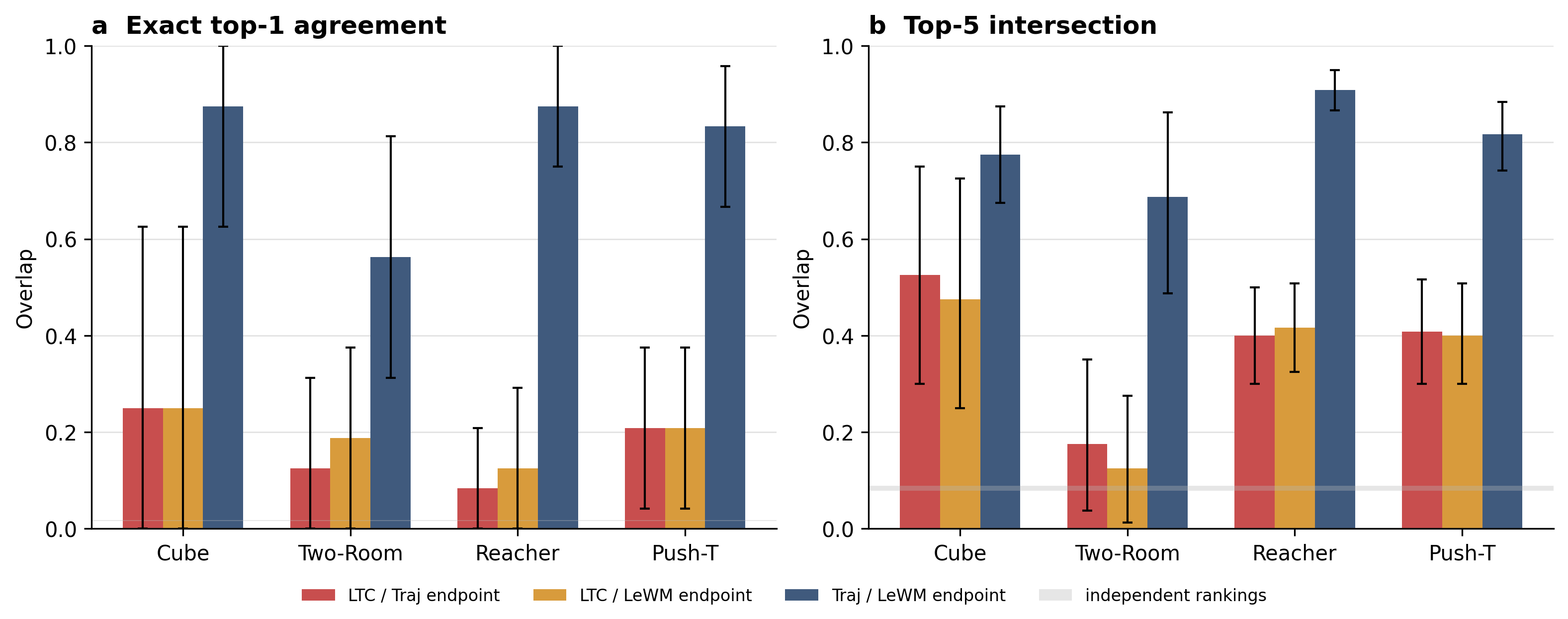}
\caption{Agreement between candidate selectors on shared candidate
pools. Panel (a) reports exact top-1 agreement, and panel (b) reports the
fraction of shared candidates in the two top-5 sets. Bars are query means,
error bars are query-level bootstrap 95\% confidence intervals, and the gray
band denotes the expectation for two independent rankings.}
\label{fig:selector-overlap}
\end{figure*}

The Traj-LeWM and LeWM endpoint selectors have exact top-1 agreement of
87.5\%, 56.3\%, 87.5\%, and 83.3\% on Cube, Two-Room, Reacher, and Push-T,
respectively; their corresponding top-5 intersections are 77.5\%, 68.8\%,
90.8\%, and 81.7\%. In contrast, LTC-versus-endpoint top-1 agreement ranges
from 8.3\% to 25.0\%, and top-5 intersection ranges from 12.5\% to 52.5\%.
This diagnostic demonstrates that LTC induces a candidate ordering substantially different from that of endpoint distance, establishing that LTC provides complementary information beyond endpoint scoring rather than serving as a mere substitute. The endpoint-matched analysis further verifies that LTC's ordering is consistent with successful execution.

\subsection{Open-Loop Latent Rollout Consistency}
\label{app:rollout-consistency}

This diagnostic asks whether LeWM or Traj-LeWM more accurately predicts future
latent states when both models receive the same recent observations and future
actions. For each task, we select one window from each of 1,500 distinct expert
episodes. Both models receive the first three observations and the identical
recorded action sequence, then autoregressively predict the next ten latent
states. At each prediction step, the corresponding observed future frame is
encoded by the model being evaluated to obtain its reference latent state.

Because latent dimensions have different numerical scales, we normalize the
squared error in each dimension by its variance, estimated from a separate set
of 500 episodes, and take the root mean across dimensions. For every episode
and prediction step, we then compute the paired difference
\[
  \Delta e=e_{\mathrm{Traj\text{-}LeWM}}-e_{\mathrm{LeWM}}.
\]
Thus, a negative value means that Traj-LeWM has lower open-loop latent
prediction error for the same input window and action sequence.

To summarize the planning-relevant range, we average $\Delta e$ over the first
five prediction steps for each episode and then average across episodes. The
result is $-0.00251$ on Push-T (95\% CI $[-0.00440,-0.00064]$),
$+0.00030$ on Cube ($[-0.00122,0.00182]$),
$-0.00343$ on Reacher ($[-0.00394,-0.00291]$), and
$-0.01269$ on Two-Room ($[-0.01535,-0.01013]$). The result is favorable on
three tasks and indistinguishable from zero on Cube.

\begin{figure*}[t]
\centering
\includegraphics[width=0.72\textwidth]{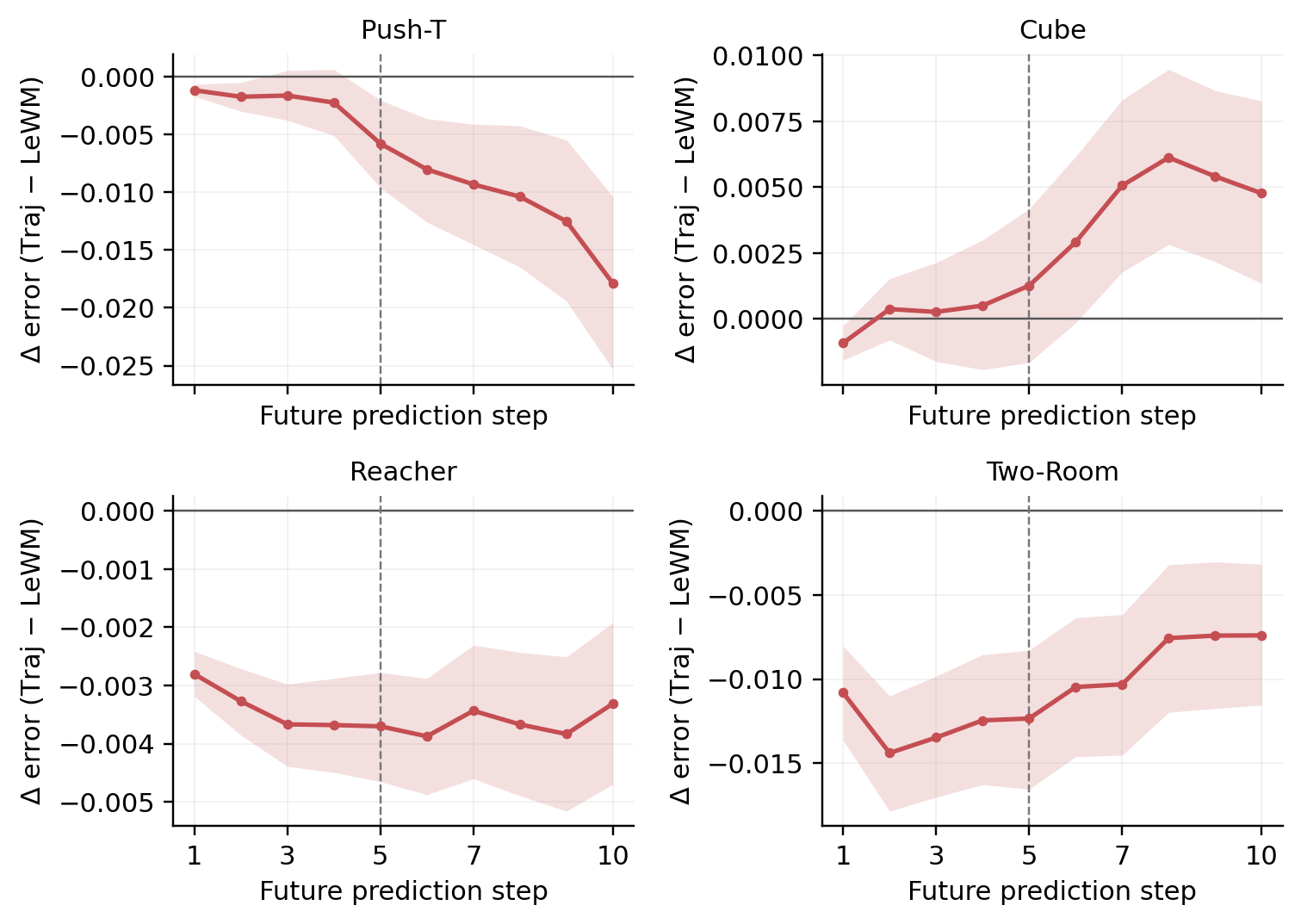}
\caption{Paired change in variance-normalized open-loop latent rollout error.
Negative values favor Traj-LeWM; shading denotes paired-bootstrap 95\%
confidence intervals. The dashed line marks the five-step planning horizon.}
\label{fig:open-loop-error}
\end{figure*}

\FloatBarrier

\section{Real-Robot Evaluation}
\label{app:real-robot-evaluation}

\subsection{Data and Paired Evaluation}
\label{app:real-robot-visuals}

\paragraph{Data.}

Experiments were conducted using a Franka FR3 robot at a sampling frequency of 30 Hz. Each frame stores an eight-dimensional state and an eight-dimensional action, comprising seven arm joints and the gripper, together with three synchronized $640\times480$ RGB streams.

\paragraph{Paired protocol.}
The supplied evaluation report contains 20 paired episode--start samples.
LeWM and Traj-LeWM receive the same initial observation and target joint
state in each pair, and each model is executed once. The target is 45 recorded
30-Hz steps after the start, equivalent to 1.5 s, 15 10-Hz endpoints, or three
0.5-s macro steps. For sample $i$ and model $m$, the report records the
smallest seven-joint distance reached during the rollout,
\begingroup
\setlength{\abovedisplayskip}{6pt}
\setlength{\abovedisplayshortskip}{6pt}
\setlength{\belowdisplayskip}{6pt}
\setlength{\belowdisplayshortskip}{6pt}
\[
  d^{\min}_{i,m}=\min_t\left\lVert q_{i,m}(t)-q_i^{\mathrm{goal}}\right\rVert_2.
\]
\endgroup
A rollout is successful only if it is not manually terminated for collision
and $d^{\min}_{i,m}\leq0.11$ rad. The threshold is the rounded Euclidean norm
of the retained per-joint tolerances
$[0.020,0.050,0.035,0.050,0.025,0.055,0.040]$ rad, whose norm is 0.108972 rad.
The real-robot sequence in Figure~\ref{fig:real-robot-dataset-views} shows five
synchronized time points from one archived episode through two external RGB
cameras and the D515 camera. These frames document the retained observation
streams only and are not associated with a LeWM or Traj-LeWM evaluation
outcome.

\medskip
\begin{center}
\centering
\includegraphics[width=\linewidth,height=0.31\textheight,keepaspectratio]{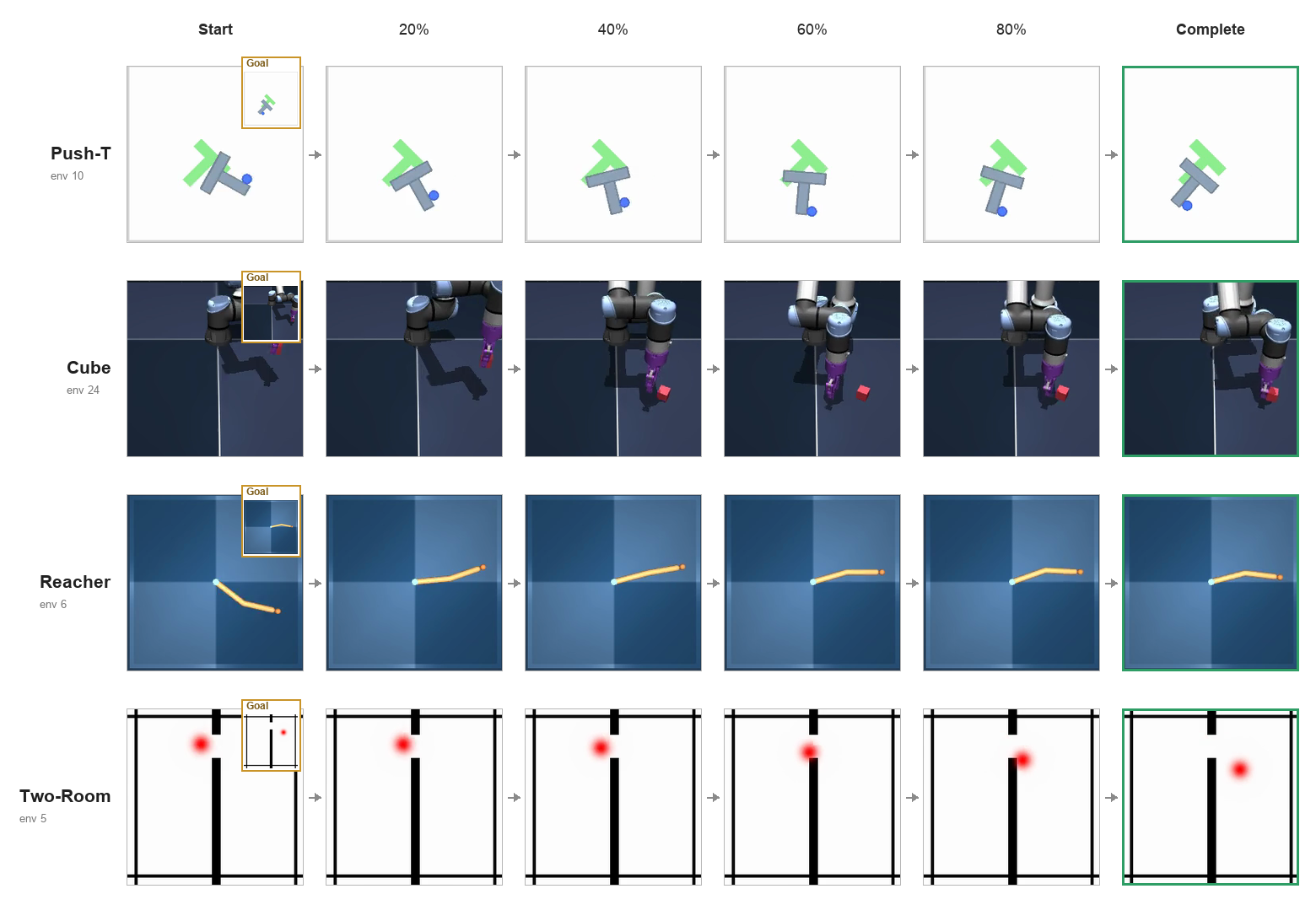}
\captionof{figure}{Representative retained Traj-LeWM rollouts for Push-T, Cube,
Reacher, and Two-Room, with each row progressing from the start observation to
the final frame.}
\label{fig:sim-rollout-gallery}
\end{center}

\vspace{0.4em}
\begin{center}
\centering
\includegraphics[width=\linewidth,height=0.24\textheight,keepaspectratio]{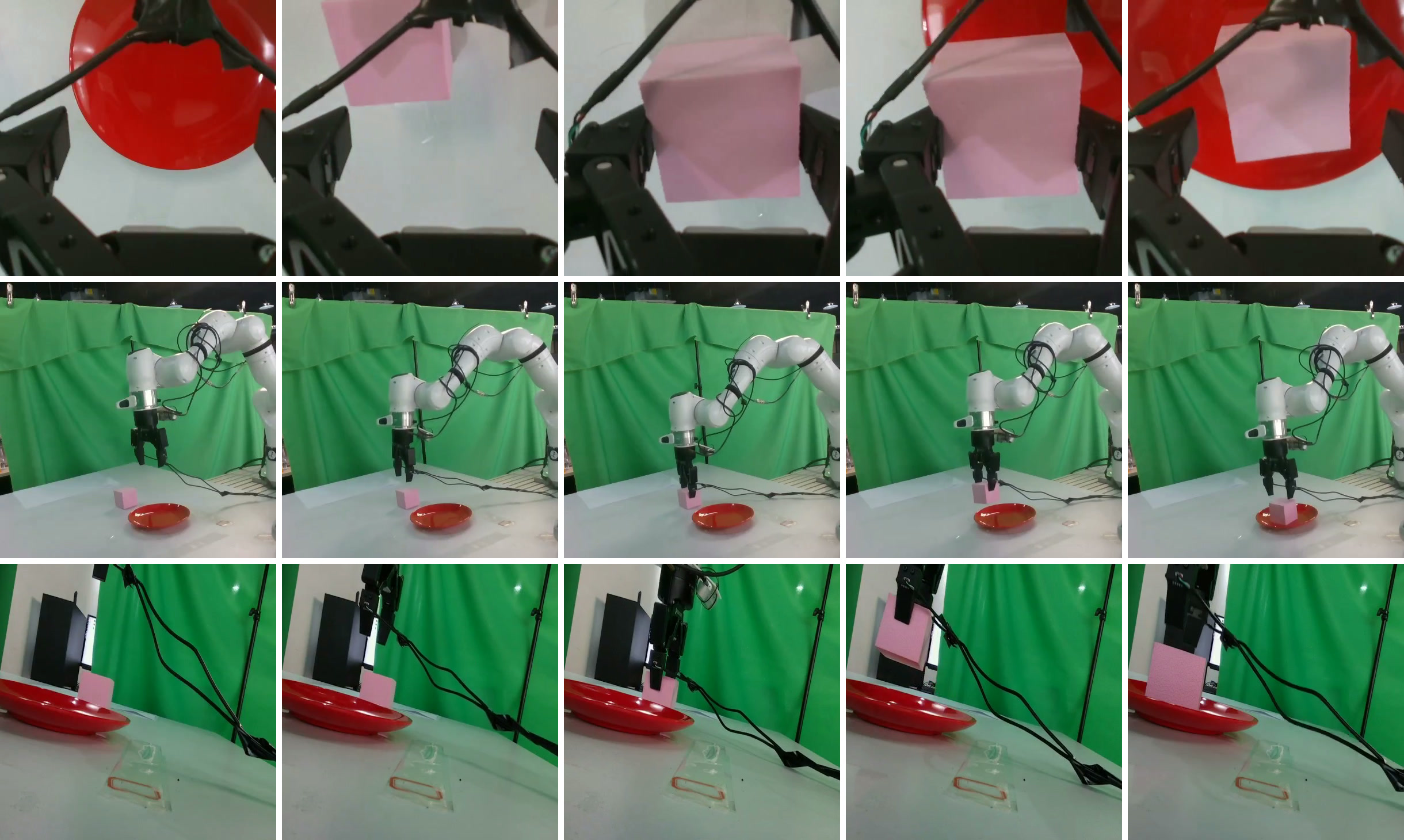}
\captionof{figure}{One Franka FR3 demonstration captured through two external cameras
and the D515 camera at five synchronized time points. These sequences document
completed executions and recorded observations.}
\label{fig:real-robot-dataset-views}
\end{center}

\FloatBarrier

\end{document}